\documentclass{INTERSPEECH2023}

\interspeechcameraready

\usepackage{arydshln}
\usepackage[scale]{tgheros}
\usepackage{lipsum}
\usepackage{abstract}
\usepackage{multirow}
\usepackage{hyperref}
\usepackage{amssymb,amsmath}

\usepackage{footnote}
\usepackage{multirow}
\usepackage{siunitx}
\usepackage{graphicx}
\usepackage{array}
\usepackage{arydshln}
\usepackage{subcaption}
\usepackage{graphicx}
\usepackage{url}
\usepackage{hyperref}

\usepackage{csquotes}
\usepackage[symbol]{footmisc}

\renewcommand{\paragraph}[1]{\noindent\textbf{#1}}

\title{Low-Resource Cross-Lingual Adaptive Training for Nigerian Pidgin}
\name{Pin-Jie Lin$^{*1,2\footnote{12}}$, Muhammed Saeed$^{*1}$, Ernie Chang$^{*3}$, Merel Scholman$^{2,4}$}
\address{
  $^1$Saarland Informatics Campus, Germany\\
  $^2$Language Science and Technology, Saarland University, Germany\\
  $^3$Reality Labs, Meta Inc.\\
  $^4$ILS, Utrecht University, the Netherlands}
\email{\{pinjie, musaeed, m.c.j.scholman\}@lst.uni-saarland.de, erniecyc@meta.com, m.c.j.scholman@coli.uni-saarland.de}

\begin{document}
\maketitle

\newcommand\blfootnote[1]{%
  \begingroup
  \renewcommand\thefootnote{}\footnote{#1}%
  \endgroup
}

\begin{abstract}

Developing effective spoken language processing systems for low-resource languages poses several challenges due to the lack of parallel data and limited resources for fine-tuning models. 
In this work, we target on improving upon both text classification and translation of Nigerian Pidgin (Naija) by collecting a large-scale parallel English-Pidgin corpus and further propose a framework of cross-lingual adaptive training that includes both continual and task adaptive training so as to adapt a base pre-trained model to low-resource languages. 
Our studies show that English pre-trained language models serve as a stronger prior than multilingual language models on English-Pidgin tasks with up to $2.38$ BLEU improvements;
and demonstrate that augmenting orthographic data and using task adaptive training with back-translation can have a significant impact on model performance. 

\end{abstract}

\section{Introduction}

\blfootnote{$\ast$ Equal contribution.}

Over the past few years, there has been an increasing interest in developing spoken language processing systems for low-resource languages such as the Nigerian Pidgin (Naija)~\cite{DBLP:journals/corr/abs-2109-06074,chang2021unsupervised}.
With a population of $75$ million people in Nigeria, Nigerian Pidgin is a low-resource language that lacks sufficient data for spoken language processing tasks. Consequently, models tend to underperform when it comes to critical tasks, such as sentiment analysis~\cite{muhammad2022naijasenti} and machine translation~\cite{pcm_paper_baseline}. 
Additionally, the orthographic variation of low-resource languages presents a challenge for language processing models, which can be addressed by collecting diverse datasets and performing data augmentation using the target language lexicon~\cite{DBLP:journals/corr/abs-2009-05460,feldman-coto-solano-2020-neural,chakravarthi2021survey}. The absence of parallel Pidgin data creates a considerable obstacle to training neural models with a high number of parameters. It also poses difficulties for fine-tuning pre-trained models on the tasks involving Pidgin language with limited resources, as seen in spoken machine translation and text classification~\cite{DBLP:journals/corr/abs-2003-12450,chang2022dialogue,alabi-etal-2022-adapting}.

In this paper, we mitigate the issues of data scarcity by collecting and releasing a large-scale parallel English-Pidgin corpus (Section \ref{sec:corpuscollection}). 
English being the lexifier of Pidgin proves to be a useful high resource language for pivoting Nigerian Pidgin to other languages~\cite{chang-etal-2022-shot}. 
Thus, we use this English-Pidgin parallel dataset to train language models. 
Prior work proposed that using multilingual models can benefit low-resource language settings~\cite{mohammadshahi2022small}. 
However, fine-tuning existing models \cite{raffel2020exploring} for specific tasks can be challenging due to their large number of parameters and sensitivity to parameter values. 
Thus, to more effectively leverage existing pre-trained models, we introduce a cross-lingual adaptive framework which involves two training procedures consisting of  
continual adaptive training and 
task adaptive training with back-translation. 
Our approach is designed to adapt a base model to a new language, making it more effective for low-resource languages.

To this end, we introduce a cross-lingual adaptation framework for fine-tuning existing models to Nigerian Pidgin~\cite{raffel2020exploring,liu-etal-2020-multilingual-denoising} (Section \ref{sec:cross_lingual_pretraining}). 
Specifically, we perform continual and task adaptation by continually pre-training language models for Naija, and then fine-tuning the models \cite{liu2019roberta,devlin2018bert} for the downstream tasks. 
In our analysis, presented in Section \ref{sec:results}, we found that the English-based model is superior to the multilingual one, indicating the importance of training on data specific to the target language. 
Additionally, we found that using task adaptive training provides a significant impact on model performance in the low-data setting. 
Our results suggest that cross-lingual adaptive training is a promising approach for building effective spoken language systems for low-resource languages\footnote{\url{https://github.com/muhammed-saeed/CLaT}.}.

Our main contributions are as follows:

\begin{itemize}
    \item We release the first large-scale English-Pidgin dataset\footnote{We release the English-Pidgin dataset and 5 million synthetic parallel corpus at \url{https://drive.google.com/file/d/1GOi0h5yU9XPZFRDYCa-hjRzf_Nx_uRE1/view}}to our knowledge, which consists of $29.73$K sentence pairs.  
    % \item Using the collected corpus, we trained a baseline machine translation model, and release a $5$ million synthetic parallel corpora generated using this system. 
    \item Using the collected corpus, we trained a baseline machine translation model, and release a corpus with $5$ million synthetic sentence pairs generated using this system. 
    We further improve upon this translation model with task adaptive training~\cite{abdulmumin2021enhanced}, and demonstrate a significant BLEU improvement of $2.28$ and $1.69$ for Pidgin-English and English-Pidgin respectively over the baseline model. 
    \item We show that the English-based pre-trained model (\textsc{T5})~\cite{raffel2020exploring} outperforms its multilingual variant (\textsc{mT5})~\cite{xue2021mt5} by $2.38$ BLEU in English-to-Pidgin translation, demonstrating the superiority of English models over multilingual one on English-Pidigin and Pidgin-English translations.
\end{itemize}

\begin{table}[t]
\small
\caption{\small \textbf{Overview of Pidgin datasets.} \textsc{En.} indicate English language and \textsc{Pg.} for Pidgin language. Data included in the corpus, along with their size in a number of sentences.
}\label{tab:data}
\centering
\begin{tabular}{l @{\hspace{0.5\tabcolsep}} c @{\hspace{0.1\tabcolsep}} r l}
\toprule
{\bfseries Corpus} & {\bfseries Language} & {\bfseries $|$Train$|$} & {\bfseries Domain} \\
\midrule
\multicolumn{4}{c}{\textsc{Parallel}} \\
\midrule
Bible & \textsc{En.}, \textsc{Pg.} & $29,737$ & religious \\
JW300 ~\cite{agic2019} & \textsc{En.}, \textsc{Pg.} & $20,218$ & religious \\
\text{Naija Treebank} ~\cite{caron2019surface}  & \textsc{En.}, \textsc{Pg.} & $9,240$ & misc. \\
\midrule
\multicolumn{4}{c}{\textsc{Monolingual}} \\
\midrule
NaijaSenti ~\cite{muhammad2022naijasenti} & \textsc{Pg.} & $8,524$ & social media \\
Afri-BERTa  ~\cite{Afri-BERT} & \textsc{Pg.} & $176,843$ & news, misc.  \\
\href{https://www.bbc.com/pidgin}{BBC Pidgin}  & \textsc{Pg.} & $4,147$ & 
 news \\ %4,147 Pidgin articles
ASR  ~\cite{ajisafe2020towards}& \textsc{Pg.} & $7,958$ & news \\
PidginUNMT ~\cite{ogueji2019pidginunmt} & \textsc{Pg.} & $5,397$ & news \\
IWSLT'15 ~\cite{luong2015stanford} &  \textsc{En.} & $143,609$ & wiki., misc. \\
WMT14-En ~\cite{luong2015stanford} & \textsc{En.} &  $4,468,840$ & news \\
\bottomrule
\end{tabular}
\end{table}

\begin{figure*}[ht]
\includegraphics[width=1.0\textwidth,height=1.0\textheight,keepaspectratio]{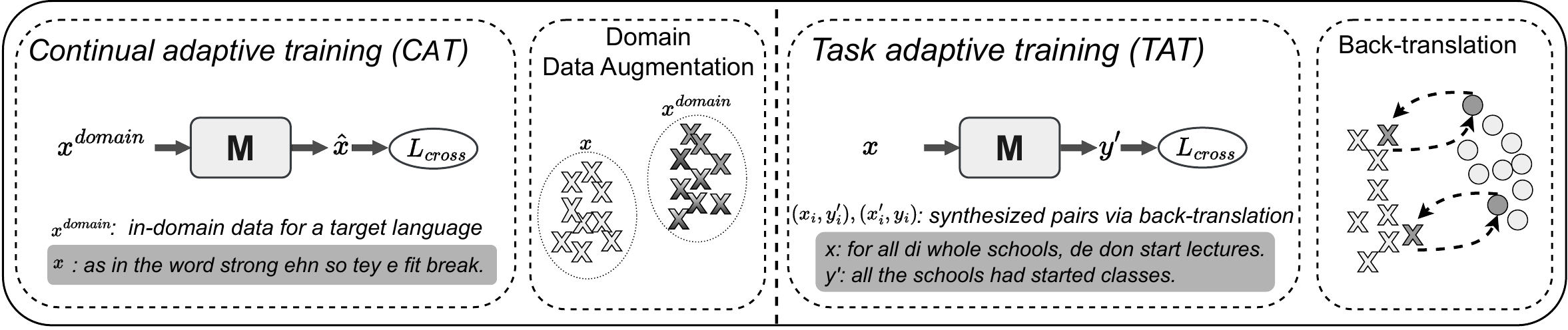}
% \includegraphics[width=1.0\textwidth,height=1.0\textheight,keepaspectratio]{LaTeX/diagrams/diagram_horizontal.pdf}
% (1) We first gather the parallel English-Pidgin corpus by synthesizing the translation of English sentences. Next, we adapt the pre-trained language model to handle
\caption{ \small Overview of the framework for low-resource sentiment classification and translation in Pidgin language: ($1$) \textbf{Continual adaptive training}: We consider a base model $M$ and a set of in-domain data $x^{domain}$ in the target language. We then train $M$ with MLM objective which enables a base model to adapt to a new language domain. ($2$) \textbf{Task adaptive training}: Starting from the observed sequence in the source language, the translation model synthesizes an inference in the target language creating the pseudo sentence pair. We construct a bi-directional back-translation by involving the forward and reverse translations. Next, the combined synthetic data serves as a supplementary task for the base model which enables the language model to adapt to more complex tasks via supervised task training.}\label{fig:architecture}
\end{figure*}
%, thereby facilitating its exposure to language adaptation

%\section{Overview} 

% Our approach is two-fold:
% (1) we describe our corpus collection process for Pidgin, which includes diverse datasets and data augmentation techniques (section \ref{}). 
% (2) We propose a novel method called Cross-Lingual Adaptive Training, which involves domain adaptive training using domain data and task adaptive training with back-translation, to adapt a base model to a new language and make it more effective for low-resource languages (section \ref{}).

\section{Corpus Collection}\label{sec:corpuscollection} 

While there have been several efforts to create datasets for Pidgin \cite{agic2019,caron2019surface}, the language still lacks a sufficiently sized dataset for application in machine translation models. 
To address this issue, we combine and enrich various parallel and monolingual texts and datasets to generate a high-quality parallel dataset. 
The Nigerian Pidgin corpus collection includes six resources: 
(1) The Holy Bible, where each verse in English was mapped to its corresponding verse in Pidgin, resulting in $29,737$ parallel sentences\footnote{We utilize the edition provided by Wycliffe Bible Translators, Inc.}. A limited number of chapters required manual processing to ensure their quality. 
(2) JW300 corpus, which contains texts from two religious magazines covering various topics. 
(3) The Naija Treebank, which is a parallel corpus of transcribed spoken Pidgin text with English translations. 
(4) The NaijaSenti corpus, which consists of $21,017$ crawled tweets in Pidgin and three additional Nigerian languages. 
(5) The Pidgin subset of the Afri-BERTa dataset, which consists of $176$K Pidgin sentences, and Pidgin text from $17$K Pidgin articles from BBC Pidgin, ASR, and PidginUNMT. 
(6) $5$ million synthetic sentence pairs in English-Pidgin, which were generated from the ISWLT'15 and WMT14 datasets and the Pidgin sentences in the monolingual corpus.
Table \ref{tab:data} presents the overview of collected datasets included in the current study along with their respective size.

\paragraph{Orthographic analysis.}\label{tab:orthovariation}
Due to the lack of a commonly accepted standard orthography in Nigerian Pidgin, we observe various forms of orthographic variation in the data. The data is characterized by both intra-textual variation (i.e.~variation within texts from the same source) and inter-textual variation (i.e.~between different sources). 
We identify four main classes of systematic variations that occur in the data: (\textsc{i}) alternation between similar sounds; (\textsc{ii}) conversion of digraphs into a single letter or alternate digraphs; (\textsc{iii}) phonetic transcription of (blended) letter pairings; and (\textsc{iv}) deletion of silent letters. Table \ref{tab:variationclass} presents examples of each of these classes.\footnote{Note that the different datasets adhere to different orthographies -- some aim to stay close to English spellings and others aim for phonetic spellings. Both the English spellings and the variations, therefore, occur in our data.}

These variations all have phonetic origins. For example, the alternation between \textquote{c} and \textquote{k} can be attributed to both consonants being ejective, and the conversion of \textquote{ee} to \textquote{i} can be attributed to both vowels having similar sounds in the Pidgin pronunciation of certain words. As such, we address the inconsistent input by collecting diverse datasets, highlighting the significance of our released data.

% original table 

\begin{table}[t]
\caption{\textbf{Types of orthographic variation in Nigerian Pidgin.}}\label{tab:variationclass}
\centering
\begin{tabular}{lll}
\toprule
\textbf{Type}          & \textbf{Subtype}       & \textbf{Example}            \\
\midrule
Alternation  & c / k         & \textbf{c}arry - \textbf{k}arry      \\
              & a / o         & c\textbf{a}ll - c\textbf{o}ll        \\
              % & y / i         & b\textbf{y} - b\textbf{i}            \\
              % & e / i         & d\textbf{e}stroy - d\textbf{i}stroy  \\ %\dataexampleshdashline
\midrule
Conversion    & ou / a        & \textbf{ou}r - \textbf{a}wa          \\
              & ou / o        & y\textbf{ou}r - y\textbf{o}r         \\
              % & au / o        & bec\textbf{au}se - bik\textbf{o}s    \\
              % & ee / i        & s\textbf{ee} - s\textbf{i}           \\
              % & ea / i        & r\textbf{ea}ch - r\textbf{i}sh       \\
              % & eo / i        & p\textbf{eo}ple - p\textbf{i}pol     \\
              % & th / d        & \textbf{th}e - \textbf{d}i           \\
              % & th / t        & \textbf{th}ing - \textbf{t}in        \\  
              % & ng / n        & thi\textbf{ng} - ti\textbf{n}        \\  
              % & ph / f        & pro\textbf{ph}et - pro\textbf{f}et    \\
              % & wh / w        & \textbf{wh}en - \textbf{w}en          \\
              % & ch / sh        & tea\textbf{ch} - ti\textbf{sh}           \\   
\midrule
Transcription & bl / bol       & trou\textbf{bl}e - tro\textbf{bol}   \\
              & er / a        & wheth\textbf{er} - wed\textbf{a}    \\               
              % & ight / ite        & n\textbf{ight} - n\textbf{ite}    \\               
\midrule
Deletion      & initial   & \textbf{h}e - e        \\
              & medial  & diff\textbf{e}rent - difren \\ 
              % & final   & com\textbf{e} - kom         \\
\bottomrule
\end{tabular}
\end{table}

\section{Cross-Lingual Adaptive Training}

\label{sec:cross_lingual_pretraining}
% To benchmark and improve the performance of low-resource Pidgin understanding and translation tasks, we introduce a two training approach which is superior to sentiment classification and neural machine translation. To overcome the orthographic variation problem, we introduce two adaptive pre-training methods before fine-tuning on downstream tasks: (1) \textsc{DaT}: \textbf{D}omain \textbf{A}daptive \textbf{T}aining where the pre-trained language model continually adapts to new language and domain. We then fine-tune a Naija language RoBERTa using only Naija corpus. 

% We aim to validate the hypothesis that transfer learning on English models performs better for languages with a close relationship to English compared to using multi-lingual models.

Considering the challenges posed by orthographic variations and the scarcity of labeled data for 
developing performant spoken language processing systems, we introduce two supplementary training approaches\textemdash adapting the model to the new language and task before fine-tuning on downstream tasks\textemdash that can be utilized to benchmark and enhance the performance of low-resource Pidgin sentiment classification and translation tasks: (1) \textsc{CaT}: \textbf{C}ontinual \textbf{A}daptive \textbf{T}aining and (2) \textsc{TaT}: \textbf{T}ask \textbf{A}daptive \textbf{T}aining.

\paragraph{Continual adaptive training.}
Given the limited availability of labeled Pidgin data, fine-tuning the large number of weights in pre-trained language models (PLMs) is challenging. To this end, we transfer the knowledge about one language absorbed in the weights to the target language by continually adapting the model to a new language via the unlabeled Pidgin corpus. The \emph{\textbf{C}ontinual \textbf{A}daptive \textbf{T}raining} (\textsc{CaT}) provides supplementary training for the base model to transfer to a specific language domain and thus improves the model's performance on the downstream task. Figure ~\ref{fig:architecture} depicts the training phase where the base model $M$ conducts language adaptation via data assuming from the same domain, thus building an adapted model specialized in a new language. More specifically, an English-based $M^{English}$ is adapted to Pidgin language using large-scale unlabeled data, resulting in a language-specific $M^{Pidgin}$. Subsequently, we fine-tuned this model for the target tasks.

% performed domain adaptation on English RoBERTa language models, as well as multi-lingual models. We chose to adapt English models because the linguistic structure of Naija is similar to English. For machine translation, we also used T5-English and MT5 (multi-lingual T5) models. The reason for this is that the T5 decoder is extensively trained in English sentences, which helps the model generate high-quality synthetic Naija-English sentence data using the Naija data as the source. Additionally, we wanted to verify our assumption that the NMT model would be able to better handle orthographic variation, as the T5 model is trained on rich data and has already been exposed to a variety of word formats. Furthermore, the fact that the T5 decoder is extensively trained in English corpora resulted in the highest-quality synthetic data for machine translation. 

\paragraph{Task adaptive training.}
To enhance the model's ability to tackle more intricate tasks, we further introduce \emph{\textbf{T}ask \textbf{A}daptive \textbf{T}raining} (\textsc{TaT}) which allows the model to adapt to the translation task through supervised learning. Our task training involves combining the two sets of synthetic data that possess shared characteristics across both source and target languages for $M$. To create synthetic data, \textsc{TaT} employs back-translation, a technique that has proven effective in low-resource machine translation scenarios. By leveraging bi-directional back-translation data in our approach, we augment the volume of task-specific training data accessible to the model which can potentially enhance the performance on more complex translation tasks. Specifically, we obtain a synthetic dataset $D'_{x{\rightarrow}y'} = \{(x,y')|x \in D\}$ via back-translation where the pseudo translation $y'$ was generated according to the sequence $x$ in the source language. We combine two translation directions as the bi-directional back-translation data $D^{BT}=D'_{x{\rightarrow}y'} \cup D'_{y{\rightarrow}x'}$.

% \emph{Bidirectional English-to-Naija machine translation}: First, using both parallel corpus and the data augmentation for orthographic variation data, we trained an existing model \cite{raffel2020exploring} for building a bi-directional NMT model. We denote the base encoder-decoder model $M$ which is pre-trained in monolingual or multilingual data. \textsc{TaT} requires pseudo translations generated by a well-trained translation model, which serve as noisy data for adapting a base translation model. 

\section{Main Results}\label{sec:results}

\paragraph{General setup.}
We closely followed the training procedure in transformers \cite{vaswani2017attention}. We trained the transformer translation models using Fairseq \cite{ott2019fairseq}. For experiments with \textsc{T5} \cite{raffel2020exploring} and \textsc{mT5} \cite{DBLP:journals/corr/abs-2010-11934}, we use Huggingface \cite{wolf-etal-2020-transformers}. We consider \textsc{Base} for all the checkpoints of the models. 
% To tackle the limited vocabulary issue we have used sentence-piece \cite{kudo2018sentencepiece}. 

\label{sec:ablation}
\subsection{Sentiment Classification}
\paragraph{Data.} We derived the low-resource dataset from NaijaSenti \cite{muhammad2022naijasenti}, which performs sentiment analysis with $3$ classes ($6.7$K/$0.6$K/$1.2$K)\footnote{We obtained the portion of the dataset from the authors.}. We report \textsc{F1} score.

\paragraph{Setup.} We leverage \textsc{RoBERTa} \cite{liu2019roberta} and \textsc{BERT} \cite{devlin2018bert} in base versions. We added \textsc{Init} baselines where the weights of models are randomly initialized and refer \emph{fine-tuning} as \textsc{FT} which directly transfers the pre-trained language model to Pidgin language. 
When performing \textsc{CaT} we continually train \textsc{RoBERTa} and \textsc{BERT} on monolingual Pidgin corpus with masked language modeling objective following the instruction in \cite{ogueji2021small}, followed by fine-tuning on multi-class classification (\textquote{positive}, \textquote{negative}, \textquote{neutral}) task.

% \paragraph{Domain adaptation for low-resource Pidgin} To perform language modeling, we use the \cite{liu2019roberta} and \cite{devlin2018bert} models trained on English corpora, as well as the multilingual BERT model. These models are adapted to the Pidgin domain by training with monolingual Pidgin data, followed by fine-tuning for multi-class classification ("positive", "negative", "neutral") using a portion of the \cite{muhammad2022naijasenti} sentiment analysis data.

% Both the RoBERTa and BERT consist of $12$ transformer blocks, $12$ attention heads, and $768$ hidden dimensions, totaling $110$M parameters. 

% Wordpiece /bpe tokenization
% we apply a fresh monolingual Pidgin-subword vocabulary and embedding was generated. For our experiment, we used both English RoBERTa and Multilingual RoBERTa to initialize all components of our model, except word embeddings. 
% For \textsc{DAT}, we employed a subword to create a subword vocabulary for our Pidgin RoBERTa. We then trained this vocabulary using a corpus made up of text from numerous sources. This created a novel monolingual Pidgin subword embedding, which only contained Pidgin subwords. While we used the WordPiece tokenizer to build the vocabulary for BERT. 

\paragraph{\textsc{CaT} improves Pidgin comprehension.}
As shown in Table ~\ref{tab:classifiaciton-resultss}, \textsc{BERT} and \textsc{RoBERTa} with continual adaptive training have both improved \text{FT} after the additional pre-training epochs on Pidgin data, resulting in  +$1$ and +$2.4$ point improvement in \textsc{F1}. Furthermore, \textsc{CaT} enables significant performance gains compared to \textsc{Init} by +$8.9$ and +$14.1$ points of \textsc{F1}. The reason for this can be attributed to poor initialization from \textsc{Init} where fine-tunes a high number of randomly initialized parameters is challenging, while pre-training and additional adaptive training enable the acquisition of a highly informative language prior to the downstream task.

% \textsc{Init} shows that directly adapting the model with randomly initialized weights to Pidgin yields worse performance compared to using pre-trained language models. This indicates that 
% We have noticed a decrease in performance for BERT, and we attribute this to the additional adaptive training, which caused BERT's weights to move significantly away from the convergence point of the target tasks.

% \usepackage{multirow}
% \usepackage{multirow}
\begin{table}[ht]
\caption{\small \textbf{Results of sentiment classification.}}
 % \textsc{Init} specifies the weights of PLM are randomly initialized and \textsc{FT} specifies model fine-tuning. \textsc{DAT} introduces additional domain adaptive pre-training on Pidgin data before fine-tuning the classification task. We trained over 100 epochs.
\label{tab:classifiaciton-resultss}
\centering
\resizebox{0.7\columnwidth}{!}{
\begin{tabular}{l|S[table-format=2.1]S[table-format=2.1]S[table-format=2.1]}
\toprule
\textbf{Model Type} & \textbf{Init} & \textbf{FT} & \textbf{\textsc{CaT}} \\
\midrule
\text{BERT} & 71.8 & 79.7 & \textbf{80.7}  \\
\text{RoBERTa} & 68.4 & 80.1 & \textbf{82.5} \\
\bottomrule
% \text{RoBERTa (Adapter)} & \multicolumn{1}{c|}{76.6} & 66 \\
% \textbf{mBERT}                        & \multicolumn{1}{c|}{}                  & \multicolumn{1}{c|}{0.697}                &                  \\ \hline
% \textbf{T5}                          & \multicolumn{1}{c|}{}                  & \multicolumn{1}{c|}{0.7178}                &                  \\  \hline
% \textbf{BART-Base }                  & \multicolumn{1}{c|}{}                  & \multicolumn{1}{c|}{0.649}           &                  \\ \hline
% \textbf{XLM-R Base}                  & \multicolumn{3}{c|}{0.733}                                                                       \\ \hline
\end{tabular}}
\end{table}

%          init    PLM   AT
% BERT             76.7  79.7
% RoBERTa   68.4   80.1 82.5

\begin{table}[ht]
\caption{ \small \textbf{Results on \textsc{JW300} translation benchmark with data augmentation (\textsc{Data Aug.}) and task adaptive training (\textsc{TaT}).}
}
\label{tab:jw300Biblevsjw300}
\centering
\resizebox{\columnwidth}{!}{
\begin{tabular}{lS[table-format=2.2]S[table-format=2.2]}
\toprule
& \textbf{English-Pidgin} & \textbf{Pidgin-English} \\ 
\midrule
\emph{Word-level} & &  \\
\midrule
\textsc{\hspace{3mm}JW300 \cite{pcm_paper_baseline}} & 17.73 & $\mathbf{24.67}$ \\
\textsc{\hspace{3mm}Data Aug.} & $\mathbf{23.87}$ & 22.61 \\
\midrule
\emph{BPE} && \\
\midrule
\textsc{\hspace{3mm}JW300 \cite{pcm_paper_baseline}} & 24.29 & 13 \\
\textsc{\hspace{3mm}Data Aug.} & 30.74 & 28.76 \\
\textsc{\hspace{3mm}Data Aug.+TaT} & $\mathbf{32.43}$ & $\mathbf{31.04}$ \\
\bottomrule
\end{tabular}}
\end{table}

%      EN-Pidgin Pidgin-EN
% JW   24.29     13
% B+J  30.74*    28.76*
% bt   32.43*    31.04

% word-level
%      EN-Pidgin Pidgin-EN
% JW   17.73    23.87
% B+J  24.67    22.61

\subsection{English-Pidgin Translation} 
% \subsection{Enabling English-Pidgin Translation} 
\paragraph{Data.} We use \textsc{JW300} translation benchmark \cite{pcm_paper_baseline}. The baseline model uses the \textsc{JW300} parallel English-Pidgin dataset only. For augmented data, we consider \textsc{Bible} which consists of $29$K \cite{pcm_paper_baseline}. All models are evaluated on the test set using BLEU score.
% \paragraph{Data}: we run the machine translation experiments using \cite{nekoto2020participatory} parallels dataset as well as our collected EN-Pidgin bible data. For the monolingual dataset, we are using data collected by \cite{ogueji2021small} as well as the monolingual dataset we have scrapped from BBC Pidgin News. 

\paragraph{Setup.}
% For apple-to-apple comparison with 
To facilitate a direct comparison with the Pidgin translation benchmark on \textsc{JW300} \cite{pcm_paper_baseline}, we use the identical model architectures for the baselines. The word-level model consists of $4$-$4$ encoder-decoder layers and $10$ heads with an embedding size of $300$, while BPE model has $6$-$6$ layers, $4$ heads, and an embedding size of $256$. We performed shared embedding and the shared vocabulary of size $4000$. We refer to \textsc{Data Augmentation} and \textsc{Data Aug.+TaT} as the model with data augmentation from \textsc{Bible} and the model conducting task training on the bi-directional noisy data via back-translation. We exploited back-translation (BT) to produce $430$K synthetic parallel sentences from our collected monolingual Pidgin data for \textsc{TaT}. We also release the generated $5$ million parallel sentences from the ISWLT15 and WMT14 datasets.

% For the word-level model, we used Transformer with 4-4 Layers, 10 heads, and embedding size of 300. For the subword-level, the models have 6-6 Layers, 4 heads, 1024 FFNN, and 256 embeddings. The "*" indicates using shared-all-embeddings for the specific result. We apply BPE tokenization.

% Following the Pidgin translation in \cite{pcm_paper_baseline}, we experiment with both word-level and subword-level tokenization methods for the translation models. We compared our model which is trained on data twice the size of their data. 

\paragraph{Data augmentation improves performance.}
Table \ref{tab:jw300Biblevsjw300} demonstrates that BPE model with data augmentation significantly improves the baselines by $6.45$ and $15.76$ BLEU points in both translation directions. For word-level models, augmentation leads to an increase in the BLEU score by $6.14$, while the score for Pidgin-to-English translation decreases by $2.06$ points. We analyzed the dataset and the model in order to uncover the reason for this decrease, and we found that the \textsc{$Bible$} dataset introduces a lot of orthographic variation when text is segmented at the word-level while BPE enables sharing more semantic units.

% Table \ref{tab:jw300Biblevsjw300} demonstrates that joint training with augmented orthographic data
% generally improves performance. BPE model with data augmentation significantly improves the baselines by $6.45$ and $15.76$ BLEU points in both translation directions. For word-level models, augmentation leads to an increase in the BLEU score by $6.14$, while the score for Pidgin-to-English translation decreases by $2.06$. We analyzed the dataset and the model in order to uncover the reason for this decrease, and we found that the \textsc{$Bible$} dataset introduces a lot of orthographic variation when text is segmented at the word-level while BPE enables sharing more semantic units. Furthermore, \textsc{TaT} further improve upon this translation model with the +$2.28$ and +$1.69$ point improvement for Pidgin-English and English-Pidgin respectively.

%Multi-lingual models translation results from the best checkpoint of 20 epochs of training, record using sacrebleu, trained on a combination of Bible, JW300, TreeBank

% \scalebox{0.96}{}
\begin{table}[ht]
\caption{ \small \textbf{Results on JW300 translation benchmark using \textsc{T5} and \textsc{mT5}.}}
\label{tab:t5}
\centering
\begin{tabular}{lS[table-format=2.2]S[table-format=2.2]}
\toprule
\textbf{Model Type} & \textbf{English-Pidgin} & \textbf{Pidgin-English} \\ 
\midrule
\emph{\textsc{JW300, Bible}} \\
% \textsc{\hspace{3mm}mT5 (\small{small})} & 32.82  & 31.37 \\
\textsc{\hspace{3mm}mT5 (\small base)} & 33.78 & 32.4  \\ 
\textsc{\hspace{3mm}T5 (base)} & \textbf{36.16} & \textbf{33.22} \\
\midrule
\emph{\textsc{All}} \\
\textsc{\hspace{3mm}mT5 (\small base)} &33.92 & 32.75  \\ 
\textsc{\hspace{3mm}T5 (base)} & \textbf{36.04} & \textbf{34.02} \\ 
\midrule
\emph{\textsc{All}+\textsc{TaT}} \\
\textsc{\hspace{3mm}T5 (base)} & \textbf{36.35} & \textbf{34.04}  \\ 
\bottomrule
% 36.04 33.99
\end{tabular}
\end{table}

% and that training a larger transformer with both JW300 a performance increase of Pidgin-to-English translation above the 24.67 reported in \cite{pcm_paper_baseline}. This is due to the addition of 29K more sentences to a model with the same number of parameters as the models in \cite{pcm_paper_baseline}.

\paragraph{\textsc{TaT} with back-translation yields further improvement.}
As the investigation of \textsc{TaT}'s effectiveness, we generated corresponding parallel sentences by using monolingual Pidgin data with the \textsc{T5} for $3$ epochs of training. Table~\ref{tab:jw300Biblevsjw300} shows that \textsc{TaT} further improve upon the translation models with the +$2.28$ and +$1.69$ BLEU improvement for Pidgin-English and English-Pidgin respectively. This indicates that task adaptive training with back-translation training provides a better initialization for machine translation tasks.

% the model results for translation from English to Pidgin were $30.74$, and for Pidgin to English were $28.76$. We then took this model and did further fine-tuning epochs using the real-parallel dataset, and the final result was $32.43$ and $31.04$. After training \textsc{mT5}  with this synthetic dataset for $3$ epochs,

% Table~\ref{tab:jw300Biblevsjw300} shows that back-translation further improves the data argument results by $1.69$ and $2.28$ points respectively. 

\subsection{Further Analysis}

\label{sec:analysis_multilingual}
\paragraph{English-based model is superior to multilingual models.} To validate the hypothesis of the transferability from the English monolingual model and multilingual counterpart for Pidgin language, we compare the \textsc{T5} where the encoder-decoder is extensively trained on English corpus and the multilingual variant \textsc{mT5} that was pre-trained on new Common Crawl datasets converting 101 languages. To ensure the fine-tuning of T5 variation models converges smoothly, we train both the base version of \textsc{T5} and \textsc{mT5} in the \textsc{Data aug.} setting using \textsc{JW300} and \textsc{Bible}. Additionally, we employed \textsc{All}  the parallel corpus, which consists of \textsc{Bible}, \textsc{JW300}, and \textsc{TreeBank}.
Table ~\ref{tab:t5} demonstrates that \textsc{T5} based solely on the English language outperforms its multilingual counterparts in various scenarios which confirms our hypothesis. 
We observed a BLEU improvement of +$2.38$ and +$2.12$ for both data settings in English-Pidgin translation, while the improvement was +$0.82$ and +$1.27$ points in Pidgin-English translation. We concluded that the English-based model is superior to the multilingual one. 
Moreover, despite using more training data during training, \textsc{TaT} still slightly improves upon \textsc{T5} baselines. Next, we delve deeper into the potential of task adaptation in improving the adaptability of the base model when faced with limited labeled data.

% We have found that the orthographic data besides the treebank, increased the English-Pidgin by $0.254$ and the Pidgin-English $0.76$ BLUE points. 

% \paragraph{Impact of orthographic data.}
% We have augmented the parallel training corpus with orthographic variation by synthetizing 1051 data points. We use \textsc{T5} \cite{raffel2020exploring} for English-Pidigin and Pidgin-English translation tasks. Furthermore, we have also tested both \textsc{T5} models trained on "\textsc{JW300}, \textsc{Bible}" and "\textsc{JW300}, \textsc{Bible}, \textsc{Treebank}", orthographic data augmentation" on \textsc{JW300} test set and we found that training more with our parallel data increased the translation performance Pidgin to English by $0.57$  while reducing the BLEU score between English to Pidgin by $0.12$ BLEU.

\begin{figure}[ht]
    \centering
    \begin{subfigure}[b]{0.43\textwidth}
       \includegraphics[width=1\linewidth]{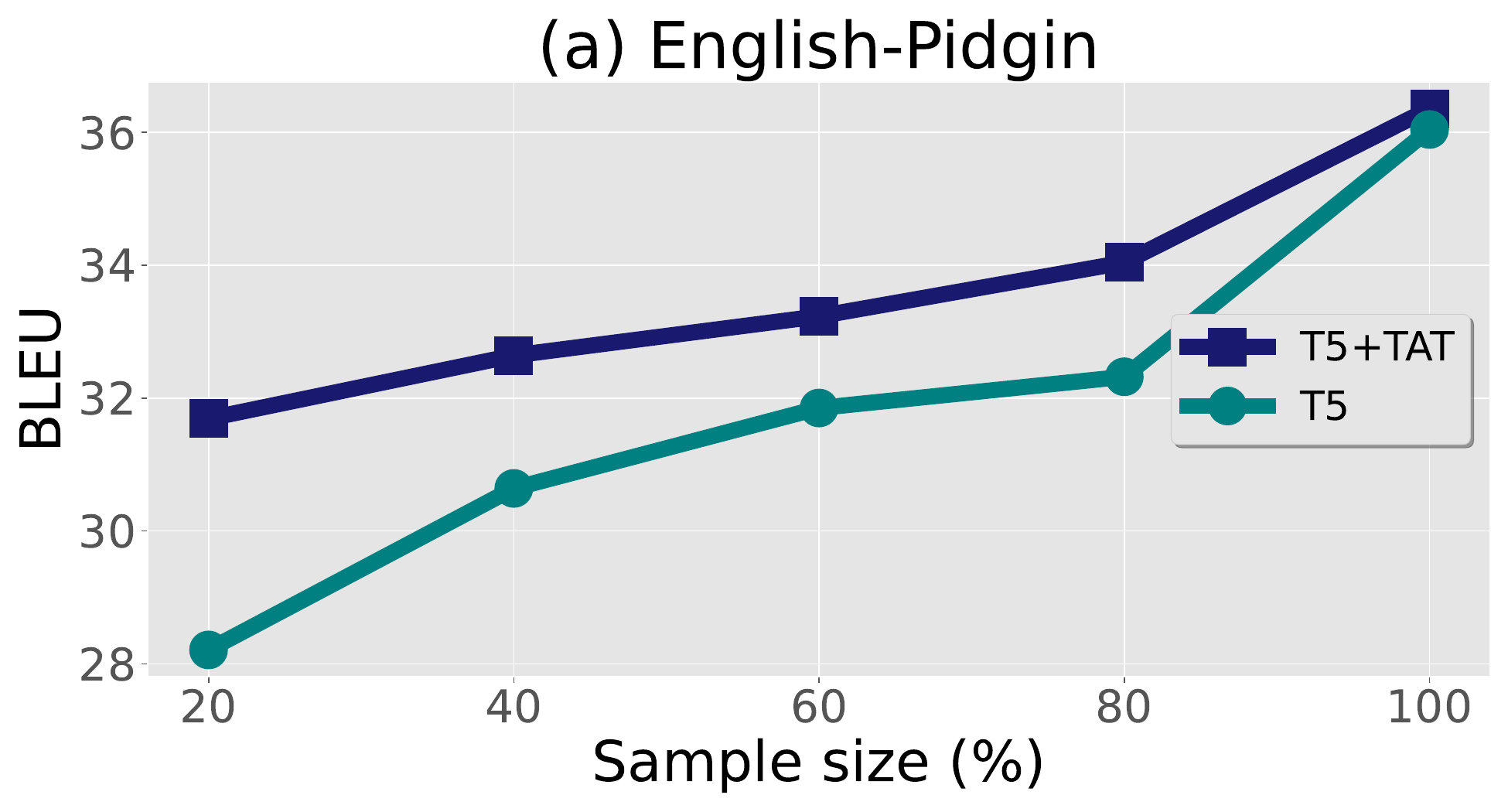}
       \label{fig:en2pcm_bleu}
    \end{subfigure}
    \begin{subfigure}[b]{0.43\textwidth}
       \includegraphics[width=1\linewidth]{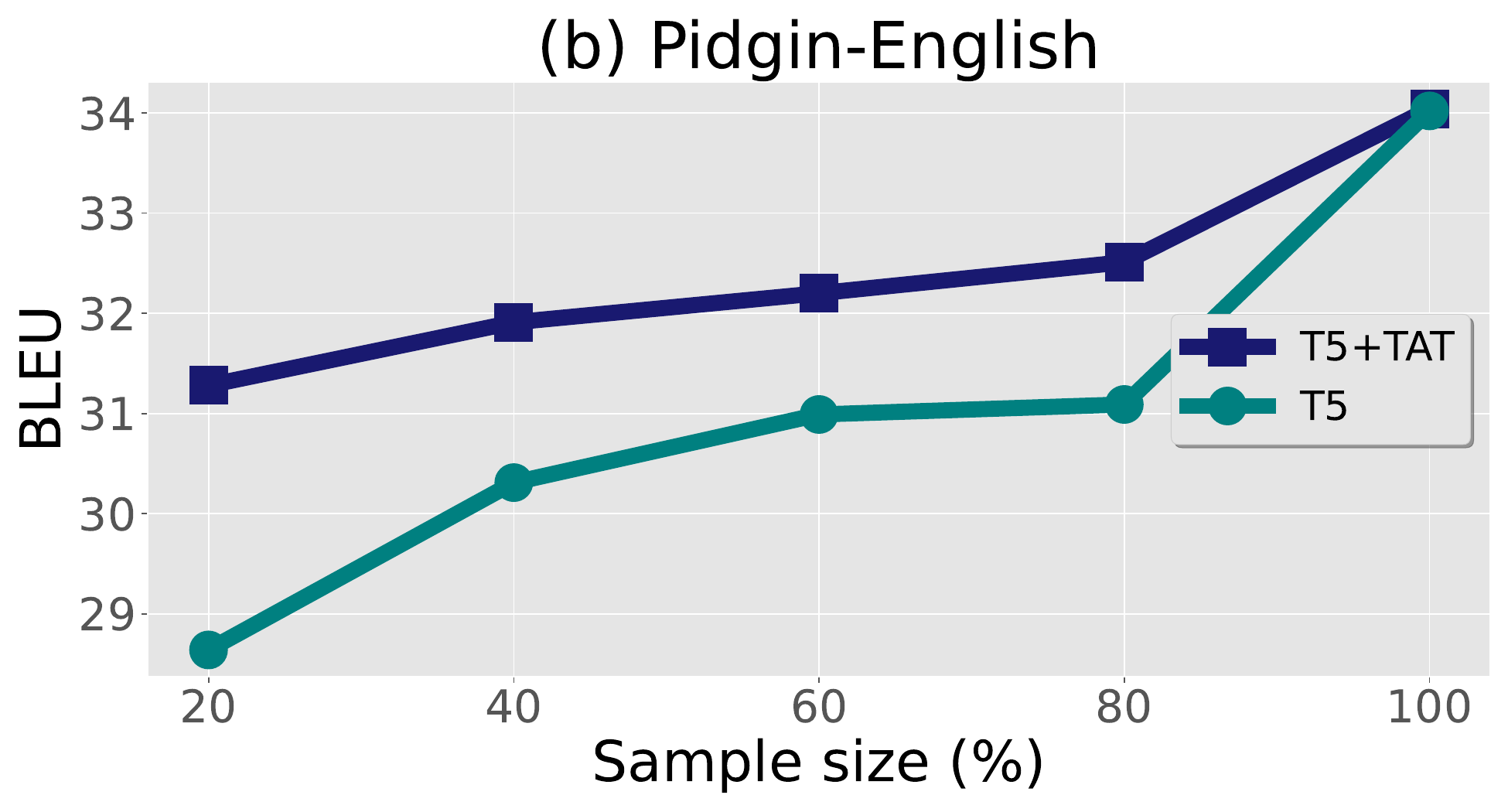}
       \label{fig:pcm2en_bleu}
    \end{subfigure}
    \caption[Two numerical solutions]{
    \small BLEU scores on $20$\%, $40$\%, $60$\%, $80$\% of sample size and full sample size of (a) English-Pidgin and (b) Pidgin-English translation tasks using T5+TAT framework.}\label{fig:low_data_translation}
\end{figure}

\paragraph{\textsc{TaT} significantly improves performance in low-data setting.}
We compare the model with task adaptation stage \textsc{T5+TaT} and the baseline \textsc{T5} to investigate the impact of task adaptation in low-data scenarios. We used four subsets randomly sampled from the original training splits (20\%, 40\%, 60\%, and 80\%) in addition to the full training set. The experimental setting was consistent with that used for English-based \textsc{T5}. Figure~\ref{fig:low_data_translation} shows that \textsc{T5+TaT} substantially outperforms the baselines across $5$ sample sizes. We observed that employing \text{TaT} obtain particularly strong performance by +$3.48$ and +$2.64$ BLEU improvement for Pidgin-English
and English-Pidgin respectively when only 20\% of the data is available for training.
Further, incorporating supervised task training into the model shows a steady increase across $5$ training splits while the performances of the baseline are sensible to the sample size.
This indicates the \textsc{T5+TaT} acquired the orthographic information from the task adaptation stage. Thus, the \textsc{T5+TaT} is capable of achieving high performance with less labeled data. The findings suggest that a robust initialization of the language model is essential for performing well in scenarios where data availability is limited, which is often the case in low-resource machine translation applications. Overall, these results highlight the potential value of incorporating \textsc{TaT} into models and suggest avenues for further research into optimizing models for limited data scenarios.

\section{Conclusion and Future Works}
In this research, we developed an effective spoken language processing framework for Pidgin language text, a low-resource language. 
We collected the largest parallel English-Pidgin corpus, performed large-scale data augmentation, and proposed a framework for cross-lingual adaptive training. Our studies show that the approach outperforms multilingual models and significantly improves model performance. Our results suggest that cross-lingual adaptive training is a promising approach for spoken language processing systems in low-resource language.
For future work, we aim to improve upon the adaptation techniques by better leveraging the English-based PLMs, and making the finetuning process more parameter-efficient for low-resource scenarios.
% Another future direction is to improve upon the synthetic data quality via more fine-grained techniques that consider the alignment information.

\section{Acknowledgements}
This work was supported by the Deutsche Forschungsgemeinschaft, Funder Id: \url{http://dx.doi.org/10.13039/501100001659}, Grant Number: SFB1102: Information Density and Linguistic Encoding.

\bibliographystyle{IEEEtran}
\bibliography{custom}

\end{document}